\documentclass[letterpaper, 10 pt, conference]{ieeeconf}

\IEEEoverridecommandlockouts  

\usepackage{graphics} % for pdf, bitmapped graphics files
\usepackage{epsfig} % for postscript graphics files
\usepackage{mathptmx} % assumes new font selection scheme installed
\usepackage{times} % assumes new font selection scheme installed
\usepackage{amsmath} % assumes amsmath package installed
\usepackage{amssymb}  % assumes amsmath package installed
\usepackage{color}
\usepackage[linesnumbered,ruled,vlined]{algorithm2e}
\usepackage{bm}

\newtheorem{definition}{Definition}

\newtheorem{prop}{Proposition}
\newtheorem{problem}{Problem}

\newtheorem{examp}{Example}

\title{\LARGE \bf Negotiating the Probabilistic Satisfaction of Temporal Logic Motion Specifications}

\author{Igor Cizelj and Calin Belta
\thanks{This work was partially supported by 
the ONR  MURI under grant N00014-10-10952
and by the NSF under grant CNS-0834260. }
\thanks{The authors are with the Division of Systems Engineering at Boston University, Boston, MA 02215, USA. Email:
     {\tt\small $\{$icizelj,cbelta$\}$@bu.edu}.}%
}

\begin{document}
\maketitle
\thispagestyle{empty}
\pagestyle{empty}

\begin{abstract}
We propose a human-supervised control synthesis method for a stochastic Dubins vehicle such that the probability of satisfying a specification given as a formula in a fragment of Probabilistic Computational Tree Logic (PCTL) over a set of environmental properties is maximized. Under some mild assumptions, we construct a finite approximation for the motion of the vehicle in the form of a tree-structured Markov Decision Process (MDP). We introduce an efficient algorithm, which exploits the tree structure of the MDP, for synthesizing a control policy that maximizes the probability of satisfaction. For the proposed PCTL fragment, we define the specification update rules that guarantee the increase (or decrease) of the satisfaction probability. We introduce an incremental algorithm for synthesizing an updated MDP control policy that reuses the initial solution. The initial specification can be updated, using the rules, until the supervisor is satisfied with both the updated specification and the corresponding satisfaction probability.  We propose an offline and an online application of this method. 
\end{abstract}

\section{Introduction} \label{introduction}

Temporal logics, such as Linear Temporal Logic (LTL) and Computational Tree Logic (CTL), have been recently employed to express complex robot behaviors such as ``go to region A and avoid region B unless regions C or D are visited'' (see, for example, \cite{kress-gazit:whereswaldo?}, \cite{karaman:vehicle}, \cite{kloetzer:fully}, \cite{Murray2009}, \cite{6016581}).

In order to use existing model checking and automata game tools for motion planning (see \cite{baier:principles}), many of the above-mentioned works rely on the assumption that the motion of the vehicle in the environment can be modeled as a finite system \cite{Clarke1999} that is either deterministic \cite{DingRecedingHorizon}, nondeterministic \cite{KlBe-HSCC08-book}, or probabilistic (\cite{LWAB10}). If a system is probabilistic, probabilistic temporal logics, such as Probabilistic CTL (PCTL) and Probabilistic LTL (PLTL), can be used for motion planning and control.
In particular, given a robot specification expressed as a probabilistic temporal logic formula, probabilistic model checking and automata game techniques can be adapted to synthesize control policies that maximize the probability that the robot satisfies the specification (\cite{LWAB10}, \cite{ProbSafeControlofNoisyDubinsVehicle}).

However, in many complex tasks, it is critically important to keep humans in the loop and engaged in the overall decision-making process. For example, during deployment, by using its local sensors, a robot might discover that some environmental properties have changed since the initial computation of the control strategy. As a result, the satisfaction probability may decrease, and the human operator should be asked whether the probability is satisfying. Alternatively, the user can change the specification according to the new environmental properties to bring the satisfaction probability over a desired threshold.  Thus, it is of great interest to investigate how humans and control synthesis algorithms can best jointly contribute to decision-making. 

To answer this question, we propose a theoretical framework for a {\emph{human-supervised control synthesis method}}. In this framework, the supervisor is relieved of low-level tasking and only specifies an initial robot specification and decides whether or not  to deploy the vehicle, based on a given specification and the corresponding satisfaction probability. The control synthesis part deals with generating control polices and the corresponding satisfaction probabilities as well as proposing updated motion specifications, to the supervisor, guaranteed to increase (or decrease) the satisfaction probability. 

We focus on controlling a stochastic version of a Dubins vehicle such that the probability of satisfying a specification given as a formula in a fragment of PCTL over a set properties at the regions in the environment is maximized. We assume that the vehicle can determine its precise initial position in a known map of the environment. However, inspired by practical applications, we assume that the vehicle is equipped with noisy actuators and, during its motion in the environment, it can only measure its angular velocity using a limited accuracy gyroscope. 
We extend our approach presented in \cite{ProbSafeControlofNoisyDubinsVehicle} to construct a finite abstraction of the  motion of the vehicle in the environment in the form of a tree-structured Markov Decision Process (MDP).
For the proposed PCTL fragment, which is rich enough to express complex motion specifications, we introduce the specification update rules that guarantee the increase (or decrease) of the satisfaction probability. 

We introduce two algorithms for synthesizing MDP control policies. The first provides an initial policy and the corresponding satisfaction probability and the second is used for obtaining an updated solution.
In general, given an MDP and a PCTL formula, solving a synthesis problem requires solving a Linear Programing (LP) problem (see \cite{baier:principles, LWAB10}). By exploiting the special tree structure of the MDP, obtained through the abstraction process, as well as the structure of the PCTL fragment, we show that our algorithms produce the optimal solution in a fast and efficient manner without solving an LP.  Moreover, the second algorithm produces an updated optimal solution by reusing the initial solution. Once the MDP control policy is obtained, by establishing a mapping between the states of the MDP and sequences of measurements obtained from the gyroscope, the policy is mapped to a vehicle feedback control strategy. We propose an offline and an online application of the method and we illustrate the method with simulations. 

The work presented in this paper is, to the best of our knowledge, novel. In ~\cite{5979895} the authors introduce the problem of automatic formula revision for LTL motion planning specifications. Namely, if a specification can not be satisfied on a particular environment, the framework returns information to the user regarding how the specification can be updated so it can become satisfiable. The presented work addresses a different but related problem; the problem of automatic formula revision for PCTL motion planning specifications. Additionally, our framework allows for noisy sensors and actuators and for environmental changes during the deployment. ~\cite{HadasProbabilistic, 6085683} address the problem of probabilistic satisfaction of specifications for robotic applications. In ~\cite{HadasProbabilistic} noisy sensors are assumed and in ~\cite{6085683} the probabilities arise from the way the car-like robot is abstracted to a finite state representation. In both cases the probability with which a temporal logic specification is satisfied is calculated. These methods differ from our work since they assume perfect actuators, whereas in our case, we relax this assumption.

The remainder of this paper is organized as follows. In Sec.~\ref{preliminaries}, we introduce the necessary notation and review some preliminary results. We formulate the problem and outline the approach in Sec.~\ref{Problem Formulation}. The construction of the MDP model is described in Sec.~\ref{construction of an mdp model}. In Sec.~\ref{pctl control policy generation} we propose two algorithms, one for generating an initial MDP control policy and the other for generating an updated MDP control policy. Case studies illustrating our method are presented in Sec.~\ref{case study}. 

%We conclude with final remarks and directions for future work in Sec.~\ref{conclusion and future work}.
%Results in this paper are stated without proof.

\section{Preliminaries} \label{preliminaries}
In this section, by following the standard notation for Markov Decision Processes (MDP) \cite{baier:principles}, we introduce a tree-structured MDP and give an informal introduction to Probabilistic Computation Tree Logic (PCTL).

\begin{definition}[Tree-Structured MDP] \label{tree structured map}
A tree-structured MDP $M$ is a tuple $(S,s_0,Act,A, P,\Pi,h)$, where 
$S$ is a finite set of states; $s_0 \in S$ is the initial state;  $Act$ is a finite set of actions; $A: S\rightarrow 2^{Act}$ is a function specifying the enabled actions at a state $s$; $P: S \times Act \times S \rightarrow [0,1]$ is a transition probability function such that 1) for all states $s \in S$ and actions $a \in A(s)$: $\sum_{s' \in S}P (s,a,s')=1$, 2) for all actions $a\notin A(s)$ and $s'\in S$, $P (s,a,s')=0$, and 3) for all states $s \in S \setminus s_0$ there exists exactly one state$-$action pair $(s',a) \in S \times A(s')$, s.t. $P(s',a,s) > 0$;
$\Pi$ is the set of propositions; and $h: S \rightarrow 2^{\Pi}$ is a function that assigns some propositions in $\Pi$ to each state of $s\in S$.
\end{definition}

In other words in a tree-structured MDP, each state has only one incoming transition, i.e., there are no cycles.
A path through a tree-structured MDP is a sequence of states that satisfies the transition probability of the MDP: $\omega=s_0s_1\ldots s_is_{i+1}\ldots$.  $\text{Path}^{fin}$ denotes the set of all finite paths.

\begin{definition}[MDP Control Policy]
\label{control policy}
A control policy $\mu$ of an MDP $M$ is a function $\mu : \text{Path}^{fin} \rightarrow Act$ that specifies the next action to be applied after every path. 
\end{definition}

Informally, Probabilistic Computational Tree Logic (PCTL) is a probabilistic extension of Computation Tree Logic (CTL) that includes the probabilistic operator $\mathcal{P}$.  
Formulas of PCTL are constructed by connecting propositions from a set $\Pi$ using Boolean operators ($\neg$ (negation), $\wedge$ (conjunction), and $\rightarrow$ (implication)), temporal operators ($\bigcirc$ (next), $\mathcal{U}$ (until)), and the probabilistic operator $\mathcal{P}$. 
For example, formula $\mathcal{P}_{max=?}[\neg \pi_3 \:  \mathcal{U}   \pi_4 ]$ asks for the maximum probability of reaching the states of an MDP satisfying $\pi_4$, without passing through states satisfying $\pi_3$.
The more complex formula $\mathcal{P}_{max=?} [\neg \pi_3 \, \mathcal{U} \, (\pi_4 \wedge \mathcal{P}_{> 0.5}[\neg \pi_3 \, \mathcal{U} \, \pi_1])$] asks for the maximum probability of eventually visiting states satisfying $\pi_4$ and then with probability greater than $0.5$ states satisfying $\pi_1$, while always avoiding states satisfying $\pi_3$.
Probabilistic model-checking tools, such as PRISM (see \cite{kwiatkowska:probabilistic}), can be used to find these probabilities. 
Simple adaptations of the model checking algorithms, such as the one presented in \cite{LWAB10}, can be used to find the corresponding control policies.

\section{Problem Formulation} \label{Problem Formulation}
%\subsection{Problem Formulation and Approach} \label{problem formulation}

In this paper, we develop a human-supervised control synthesis method, with an offline and online phase. In the \emph{offline phase} (i.e., before the deployment) the supervisor gives an initial  specification and the control synthesis algorithm returns the initial satisfaction probability. If the supervisor is not satisfied with the satisfaction probability, the system generates a set of specification relaxations that guarantee an increase in the satisfaction probability. The offline phase ends when the supervisor agrees with a specification and the corresponding satisfaction probability. 

In the {\emph{online phase}} (i.e., during the deployment), events occurring  in the environment  can affect the satisfaction probability. If such an event occurs, the system returns the updated control policy, and if necessary (i.e., if the probability decreases) proposes an updated specification that will increase the satisfaction probability. At the end of a negotiation process similar to the one described above, the supervisor agrees with one of the options recommended by the system. While the robot is stopped during the negotiation process, it is necessary that the time required for recomputing the policies be short. 

\subsection{Models and specifications}
\emph{Motion model:} A Dubins vehicle (\cite{Dubins:Oncurves}) is a unicycle with constant forward speed and bounded turning radius moving in a plane. In this paper, we consider a stochastic version of a Dubins vehicle, which captures actuator noise:
\begin{equation}
\label{dubins kinematics}
\begin{bmatrix}
 \dot x\\  \dot y\\ \dot \theta
\end{bmatrix}
= 
\begin{bmatrix}
\cos(\theta)\\ 
\sin(\theta)\\ 
u+\epsilon
\end{bmatrix}, \text{ } u  \in U,
\end{equation}
where $(x,y) \in \mathbb{R}^2$ and $\theta \in [0,2 \pi)$ are the position and orientation of the vehicle in a world frame, 
$u$ is the control input (angular velocity before being corrupted by noise), $U$ is the control constraint set, and $\epsilon$ is a random variable modeling the actuator noise. For simplicity, we assume that $\epsilon$ is uniformly distributed on the bounded interval $[-\epsilon_{max},\epsilon_{max}]$. However, our approach works for any  continuous probability distribution supported on a bounded interval. 
The forward speed is normalized to $1$.
We denote the state of the system by $q=[x,y,\theta]^T \in SE(2)$.

Motivated by the fact that the optimal Dubins paths use only three inputs (\cite{Dubins:Oncurves}), we assume $U = \{-1/\rho,0,1/\rho\}$, where $\rho$ is the minimum turn radius.
We define $$W=\{u+\epsilon | u \in U, \epsilon \in [-\epsilon_{max},\epsilon_{max}] \}$$
as the set of applied control inputs, i.e, the set of angular velocities that are applied to the system in the presence of noise. We assume that time is uniformly discretized (partitioned) into stages (intervals) of length $\Delta t$, where stage $k$ is from $(k-1)\Delta t$ to $k \Delta t$. The duration of the motion is finite and it is denoted by  $K \Delta t$.\footnote{Since PCTL has infinite time semantics, we implicitly assume after $K \Delta t$ the system remains in the state achieved at $K \Delta t$.} We denote the control input and the applied control input at stage $k$ as $u_k \in U$ and $w_k \in W$, respectively. 

We assume that the noise $\epsilon$ is piece-wise constant, i.e, it can only change at the beginning of a stage. This assumption is motivated by practical applications, in which a servo motor is used as an actuator for the turning angle (see e.g., \cite{Mazo04robustarea}). This implies that the applied control is also piece-wise constant, i.e., $w:[(k-1)\Delta t,k\Delta t] \rightarrow W$, $k=1,\ldots,K$, is constant over each stage. 

{\emph{Sensing model:}} We assume that the vehicle is equipped with only one sensor, which is a limited accuracy gyroscope.  At stage $k$, the gyroscope returns the measured interval $[\underline{w}_k,\overline{w}_k] \subset [u_k-\epsilon_{max},u_k+\epsilon_{max}]$ containing the applied control input. Motivated by practical applications, we assume that the measurement resolution of the gyroscope, i.e., the length of $[\underline{w}_k,\overline{w}_k]$, is constant, and we denote it by $\Delta \epsilon$. For simplicity of presentation, we also assume that $n \Delta \epsilon = 2 \epsilon_{max}$, for some $n \in \mathbb{Z}^+$. Then, $[-\epsilon_{max},\epsilon_{max}]$ can be partitioned\footnote{Throughout the paper, we relax the notion of a partition by allowing the endpoints of the intervals to overlap.} into $n$ intervals: $[\underline{\epsilon}_i, \overline{\epsilon}_i]$, $i=1,\ldots,n$. We denote the set of all noise intervals as $\mathcal{E} = \{[\underline{\epsilon}_1, \overline{\epsilon}_1], \ldots, [\underline{\epsilon}_n, \overline{\epsilon}_n]\}$. At stage $k$, if the applied control input is $u_k+\epsilon$, the gyroscope will return the measured interval $[\underline{w}_k,\overline{w}_k] = [u_k-\underline{\epsilon}, u_k+\overline{\epsilon}],$ where $\epsilon \in [\underline{\epsilon},\overline{\epsilon}] \in \mathcal{E}$. Since $\epsilon$ is uniformly distributed: 
\begin{equation} \label{uniform distribution}
\text{Pr}(u_k+\epsilon \in [u_k-\underline{\epsilon}_i, u_k+\overline{\epsilon}_i])=\text{Pr}(\epsilon \in [\underline{\epsilon}_i,\overline{\epsilon}_i])=\frac{1}{n}, 
\end{equation}
$[\underline{\epsilon}_i,\overline{\epsilon}_i] \in \mathcal{E}$, $i=1,\ldots,n$.

{\emph{Environment model and specification:}} The vehicle moves in a static environment $X \subseteq \mathbb{R}^2$ in which regions of interest are present. Let $\Pi$ be a finite set of propositions satisfied at the regions in the environment. Let $[\cdot] : 2^{\Pi} \rightarrow 2^X$ be a map such that $[\Theta]$, $\Theta \in 2^{\Pi}$, is the set of all positions in $X$ satisfying all and only propositions $\pi \in \Theta$. 
Inspired by a realistic scenario of an indoor vehicle leaving its charging station, we assume that the vehicle can precisely determine its initial state $q_{init}=[x_{init},y_{init},\theta_{init}]^T$ in a known map of the environment. Specification: In this work, we assume that the vehicle needs to carry out a motion specification expressed as a PCTL formula $\phi$ over $\Pi$: 
\begin{equation}
\label{general formula}
\begin{split}
\phi: &= \mathcal{P}_{max=?} [\mathcal{P}_{\geq p_1}[\varphi_1 \mathcal{U} ( \psi_1 \wedge \mathcal{P}_{\geq p_2}[\varphi_2 \mathcal{U} ( \psi_2 \wedge\\
 &\ldots \wedge \mathcal{P}_{\geq p_f}[\varphi_f \mathcal{U} \psi_f])])]],
 \end{split}
 \end{equation}
 $f \in \mathbb{Z}^+$, where $\forall j \in \{1,\ldots,f\}$, $\varphi_j$ and $\psi_j$ are PCTL formulas constructed by connecting properties from a set of propositions $\Pi$ using only Boolean operators in Conjunctive Normal Form (CNF) and Disjunctive Normal Form (DNF)\footnote{A formula is CNF if it is a conjunction of clauses, where a clause is a disjunction of propositions. A formula is in DNF if it is a disjunction of clauses, where a clause is a conjunction of propositions.}, respectively, and $p_j \in [0,1]$. We assume that $\phi$ is in Negation Normal Form (NNF), i.e., Boolean operator $\neg$ appears only in front of the propositions. 
 In order to better explain the different steps in our framework, we consider throughout the paper the following example.

\begin{figure}[h]
\center
\includegraphics[width=0.5\textwidth]{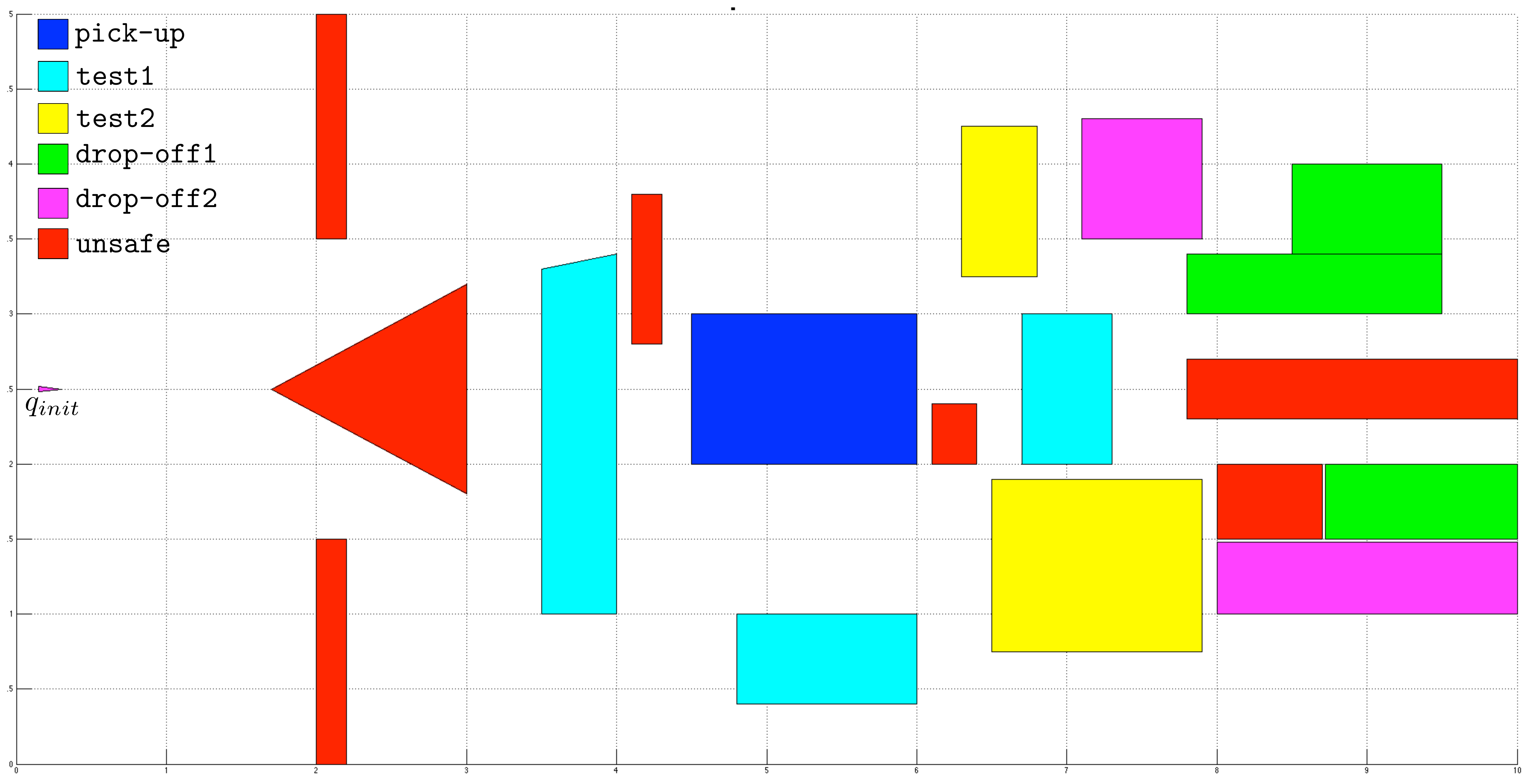}
\caption{An example and regions of interest. }
\label{example environment}
\end{figure}

\begin{examp}
Consider the environment shown in Fig.~\ref{example environment}. Let $\Pi=\{{\color{blue}\pi_p}, {\color{cyan}\pi_{t1}}, {\color{yellow}\pi_{t2}}, {\color{green}\pi_{d1}}, {\color{magenta}\pi_{d2}}, {\color{red}\pi_u}\}$, where ${\color{blue}\pi_p}, {\color{cyan}\pi_{t1}}, {\color{yellow}\pi_{t2}}, {\color{green}\pi_{d1}}, {\color{magenta}\pi_{d2}}, {\color{red}\pi_u}$ label {\ttfamily{pick-up}}, {\ttfamily{test1}}, {\ttfamily{test2}}, {\ttfamily{drop-off1}}, {\ttfamily{drop-off2}} and the {\ttfamily{unsafe}} regions, respectively. Consider the following motion specification: 

{\emph{Specification 1:}} {\emph{Starting form an initial state $q_{init}$ reach a {\ttfamily{pick-up}} region, while avoiding the {\ttfamily{test1}} regions,  to pick up a load. Then, reach a {\ttfamily{test1}} region or a {\ttfamily{test2}} region. Finally, reach a {\ttfamily{drop-off1}} or a {\ttfamily{drop-off2 }}region to drop off the load. Always avoid the {\ttfamily{unsafe}} regions}}.

The specification translates to PCTL formula $\phi$:
\begin{equation}
\label{sample formula 1}
\begin{split}
\phi: &= \mathcal{P}_{max=?}[\mathcal{P}_{>0}[\neg {\color{red}\pi_u} \wedge \neg {\color{cyan}{\pi_{t1}}} \mathcal{U} (\neg {\color{red}\pi_u} \wedge {\color{blue}\pi_p} \wedge \\ 
& \mathcal{P}_{> 0}[\neg {\color{red}\pi_u}  \mathcal{U} ((\neg {\color{red}\pi_u} \wedge {\color{cyan}\pi_{t1}}) \vee (\neg {\color{red}\pi_u} \wedge {\color{yellow}\pi_{t2}}) \wedge \\
&\mathcal{P}_{>0}[{\neg \color{red}\pi_u} \mathcal{U}  {(\neg \color{red}\pi_u} \wedge {\color{green}\pi_{d1}})  \vee (\neg {\color{red}\pi_u}  \wedge{\color{magenta}\pi_{d2}})])])]]. \text{  }  \blacksquare
\end{split}
\end{equation}
\end{examp}

Note that the proposed PCTL fragment (Eqn. (\ref{general formula})) can capture the usual properties of interest: reachability while avoiding regions and sequencing (see~\cite{fainekos2009temporal}). For example, the formula $\phi: = \mathcal{P}_{max=?}[\neg {\color{red}\pi_u} \wedge \neg {\color{cyan}{\pi_{t1}}} \wedge \neg {\color{yellow}\pi_{t2}} \mathcal{U}{\color{blue}\pi_p}]$ asks for the maximum probability of avoiding the {\ttfamily{unsafe}}, {\ttfamily{test1}} and the {\ttfamily{test2}} regions until a {\ttfamily{pick-up}} region is reached. The formula 
$\phi: = \mathcal{P}_{max=?}[\neg {\color{red}\pi_u} \wedge  \neg {\color{cyan}\pi_{t1}} \wedge \neg  {\color{yellow}\pi_{t2}} \wedge \neg  {\color{green}\pi_{d1}}  \wedge \neg {\color{magenta}\pi_{d2}} \mathcal{U} ( {\color{blue}\pi_p}  \wedge \mathcal{P}_{> 0} [\neg {\color{red}\pi_u} \wedge  \neg {\color{blue}\pi_p} \wedge \neg {\color{yellow}\pi_{t2}} \wedge \neg  {\color{green}\pi_{d1}} \wedge \neg  {\color{magenta}\pi_{d2}} \mathcal{U} ( {\color{cyan}\pi_{t1}} \wedge \mathcal{P}_{> 0} [\neg {\color{red}\pi_u} \wedge  \neg {\color{blue}\pi_p} \wedge \neg  {\color{blue}\pi_p} \wedge \neg {\color{yellow}\pi_{t2}} \wedge \neg  {\color{magenta}\pi_{d2}} \mathcal{U}  {\color{green}\pi_{d1}} ])])]$ asks for the maximum probability ov visiting a {\ttfamily{pick-up}}, {\ttfamily{test1}} and a {\ttfamily{drop-off1}} region in that order.

Next, we define the satisfaction of $\phi$ (Eqn.~\ref{general formula})  by a trajectory $q:[0,K\Delta t] \rightarrow SE(2)$ of the system from Eqn.~(\ref{dubins kinematics}). 
The word corresponding to a state trajectory $q(t)$ is a sequence $o=o_1o_2o_3\dots$, $o_k \in 2^{\Pi}$, $k \geq1$, generated 
according to the following rules, for all $t \in [0,K \Delta t]$ and $k \in \mathbb{N}$, $k \geq 1$: 1) $(x(0),y(0)) \in [o_1]$; 2) if $(x(t),y(t)) \in [o_k]$ and $o_k \neq o_{k+1}$, then $\exists$ $t' \geq t$ s.t. a) $(x(t'),y(t')) \in [o_{k+1}]$ and b) $(x(\tau),y(\tau)) \notin [\pi]$, $\forall \tau \in [t,t']$, $\forall \pi \in \Pi \setminus (o_k \cup o_{k+1})$; 3) if $(x(K \Delta t),y(K \Delta t)) \in [o_k]$ then $o_i=o_k$ $\forall i \geq k$. 
Informally, the word produced by $q(t)$ is the sequence of sets of propositions satisfied by the position $(x(t), y(t))$ of the robot as time evolves. A trajectory $q(t)$ satisfies PCTL formula $\phi$ iff the corresponding sequence satisfies the formula.

As time evolves and  a sequence $o$ is generated, we can check what part of $\phi$ is satisfied so far. If $\mathcal{P}_{\geq p_1}[\varphi_1 \mathcal{U} ( \psi_1 \wedge \ldots \wedge \mathcal{P}_{\geq p_i}[\varphi_i \mathcal{U} \psi_i]]$ part of $\phi$ is satisfied we say $\phi$ is satisfied up to $i$, $0 \leq i \leq f$ (for more details see Sec.~\ref{online control policy}).

Assume that at $k\Delta t$, for some $k=0,\ldots, K-1$, %\footnote{To accommodate the offline application of the method, the specification can be updated at $k=0$, before the vehicle is deployed.} 
 the motion specification is updated. Then, given $\phi$ satisfied up to $i$, $0 \leq i \leq f$, the updated PCTL formula, denoted $\phi^+$, is obtained from $\phi$ by removing the already satisfied part of $\phi$, and then by 1) adding or removing conjunction clause from $\psi_j$, or 2) adding or removing a disjunction clause from $\varphi_j$, or 3) increasing or decreasing $p_j$, for any $j \in \{i, \ldots, f\}$.
%\begin{itemize}
%\item removing the already satisfied part of $\phi$, and 
%\item 1) adding or removing conjunction claus from $\psi_j$, or 2) adding or removing a disjunction claus from $\varphi_j$, or 3) increasing or decreasing $p_j$, for any $j \in \{i, \ldots, f\}$.
%\end{itemize}
Formal definitions are given in Sec.~\ref{online control policy}.  To illustrate this idea consider the following example: 
\begin{examp}
Consider Specification 1 and assume that at $k\Delta t$ the vehicle enters a {\ttfamily{pick-up}} region, while avoiding the {\ttfamily{test1}} and the {\ttfamily{unsafe}} regions, and additionally, that the {\ttfamily{drop-off2}} regions become unavailable for the drop off, i.e., the vehicle is allowed to drop off the load only at the {\ttfamily{drop-off1}} regions. Then, the updated formula is:
\begin{equation*}
%\label{sample formula 2}
\begin{split}
\phi^+: &= \mathcal{P}_{max=?}[ \mathcal{P}_{> 0}[\neg {\color{red}\pi_u}  \mathcal{U} ((\neg {\color{red}\pi_u} \wedge {\color{cyan}\pi_{t1}}) \vee (\neg {\color{red}\pi_u} \wedge {\color{yellow}\pi_{t2}}) \wedge \\
&\mathcal{P}_{>0}[{\neg \color{red}\pi_u} \mathcal{U} {( \neg \color{red}\pi_u} \wedge {\color{green}\pi_{d1}})])]],
\end{split}
\end{equation*}
\end{examp}
where $\phi^+$ is obtained from $\phi$ by removing the already satisfied part of $\phi$, $\mathcal{P}_{>0}[\neg {\color{red}\pi_u} \wedge \neg {\color{cyan}{\pi_{t1}}} \mathcal{U} \neg {\color{red}\pi_u} \wedge {\color{blue}\pi_p}]$, and by removing the conjunction clause,  $(\neg {\color{red}\pi_u}  \wedge{\color{magenta}\pi_{d2}})$, from $\psi_3$. $\blacksquare$

While the vehicle moves, gyroscope measurements $[\underline{w}_k,\overline{w}_k] $ are available at each stage $k$. We define a {\it vehicle control strategy} as a map that takes as input a sequence of measured intervals $[\underline{w}_1,\overline{w}_1][\underline{w}_2,\overline{w}_2]\ldots[\underline{w}_{k-1},\overline{w}_{k-1}]$ 
and returns the control input $u_k \in U$ at stage $k$. 

\subsection{Problem formulation and approach} \label{problem formulation and approach}
We are ready to formulate the main problem that we consider in this paper:
\begin{problem}\label{problem}
Given a set of regions of interest in environment $X \subseteq \mathbb{R}^2$ satisfying propositions from set $\Pi$, a vehicle model described by Eqn.~(\ref{dubins kinematics}) with initial state $q_{init}$, an initial and updated motion specifications, expressed as PCTL formulas $\phi$ and $\phi^+$, respectively, over $\Pi$ (Eqn.~(\ref{general formula})),  find a vehicle control strategy that maximizes the probability of satisfying $\phi$ and then $\phi^+$.
\end{problem}

%\subsection{Approach} \label{approach}

Our approach to Problem \ref{problem} can be summarized as follows.
 We start by using the abstraction method presented in \cite{ProbSafeControlofNoisyDubinsVehicle} as follows: by discretizing the noise interval, we define a finite subset of the set of possible applied control inputs. We use this to define a Quantized System (QS) that approximates the original system given by Eqn.~(\ref{dubins kinematics}). Next, we capture the uncertainty in the position of the vehicle and map QS to a tree-structured MDP. Then, we develop an efficient algorithm, which exploits the tree structure of the MDP,  for obtaining an initial control policy that maximizes the probability of satisfying the initial specification. Next,  for the PCTL formulas given by Eqn.~(\ref{general formula})  we introduce the specification update rules that guarantee the increase (or decrease) of the satisfaction probability and we develop an efficient algorithm for obtaining an updated control policy, which exploits the MDP structure, the structure of the PCTL formulas (Eqn.~(\ref{general formula})), and reuses  the initial control policy.  From \cite{ProbSafeControlofNoisyDubinsVehicle} it follows that each control policy can be mapped to a vehicle control strategy and that the probability that the vehicle satisfies the corresponding specification in the original environment is bounded from below by the maximum probability of satisfying the specification on the MDP.

\section{Construction of an MDP Model} \label{construction of an mdp model}

The fact that we have introduced the initial PCTL formula $\phi$ (Eq.~(\ref{general formula})) in NNF enables us to classify the propositions in $\phi$ according to whether they represent regions that must be reached (no negation in from of the proposition) or avoided (a negation operator appears in from of the proposition). 

The abstraction process from \cite{ProbSafeControlofNoisyDubinsVehicle} can only deal with PCTL formulas where the propositions are classified into two nonintersecting sets according to whether they represent regions that must be reached or avoided. In this paper, we do not make this limiting assumption. For example, consider the PCTL formula given by Eqn.~(\ref{sample formula 1}) where the {\ttfamily{test1}} regions (i.e., proposition ${\color{cyan}\pi_{t1}}$) need to be both avoided and reached. 

\subsection{PCTL formula transformation} \label{pctl formula transformation}

In order to use the method presented in \cite{ProbSafeControlofNoisyDubinsVehicle}, we start by removing any negation operators that appear in the initial formula. To do so we use the approach presented in~\cite{fainekos2009temporal} as follows. We introduce the extended set of propositions $\Xi_{\Pi}$. In detail, we first define two new sets of symbols $\Xi_{\Pi}^+= \{ \xi_\pi | \pi \in \Pi\}$ and $\Xi_{\Pi}^-= \{ \xi_{\neg \pi} | \pi \in \Pi\}$. Then, we set $\Xi_{\Pi} = \Xi_{\Pi}^+ \cup \Xi_{\Pi}^-$. We also define a translation function {\bf{pos}}$(\phi) : \phi_{\Pi} \rightarrow \phi_{\Xi_{\Pi}}$  which takes as input a PCTL formula $\phi$ in NNF and it returns a formula {\bf{pos}}$(\phi)$ where the occurrences of terms $\pi$ and $\neg \pi$ have been replaced by the members $\xi_{\pi}$ and $\xi_{\neg \pi}$ of $\Xi_{\Pi}$ respectively. Since we have a new set of propositions, $\Xi_{\Pi}$, we need to define a new map $[\cdot]^{\Xi_{\Pi}}: \Xi_{\Pi} \rightarrow 2^X$  for the interpretation of the propositions. This is straightforward: $\forall \xi \in \Xi_{\Pi}$, if $\xi = \xi_{\pi}$ then $[\xi]^{\Xi_{\Pi}} = [\pi]$, else (i.e., if $\xi = \xi_{\neg \pi}$) $[\xi]^{\Xi_{\Pi}}= X \setminus [\pi]$ (for more details see Fig.~\ref{mdp example}). 

It can easily be seen that given a formula $\phi \in \Phi_{\Pi}$, a map $[\cdot] : \Pi \rightarrow 2^X$ and a trajectory $q(t)$ of the system from Eqn.~(\ref{dubins kinematics}), the following holds: $q(t)$ satisfies $\phi$ iff $q(t)$ satisfies {\bf{pos}}$(\phi)$. Thus, since $\phi \in \Phi_{\Pi}$ is equivalent to the formula ${\bm {\phi}}=${\bf{pos}}$(\phi)$ under the maps $[\cdot] : \Pi \rightarrow 2^X$ and $[\cdot]^{\Xi_{\Pi}} : {\Xi_{\Pi}} \rightarrow 2^X$, next results are given with respect to a formula ${\bm{\phi}} \in \Phi_{\Xi_{\Pi}}$ and a map $[\cdot]^{\Xi_{\Pi}} : {\Xi_{\Pi}} \rightarrow 2^X$. We denote all PCTL formulas in NNF without any negation operator using bold Greek letters, e.g., ${\bm \phi}$, ${\bm \phi}'$, ${\bm \phi}'_1$.

At this point we have distinguished the regions that must be avoided ($\Xi_{\Pi}^-$) and the regions that must be reached ($\Xi_{\Pi}^+$). %We proceed with the construction of the MDP. 

\subsection{Approximation} \label{approximation}

We use $q_k(t)$ and $w_k$, $t \in [(k-1)\Delta t, k\Delta t]$, $k=1,\ldots,K$ to denote the state trajectory and the constant applied control at stage $k$, respectively. 
With a slight abuse of notation, we use $q_k$ to denote the end of state trajectory $q_k(t)$, i.e., $q_k=q_k(k \Delta t)$. Given a state $q_{k-1}$, the state trajectory $q_k(t)$ can be derived by integrating the system given by Eqn.~(\ref{dubins kinematics}) from the initial state $q_{k-1}$, and taking into account that the applied control is constant and equal to $w_k$. 
Throughout the paper, we will also denote this trajectory by $q_k(q_{k-1},w_k,t)$, when we want to explicitly capture the initial state $q_{k-1}$ and the constant applied control $w_k$. 

For each interval in $\mathcal{E}$ we define a representative value $\epsilon_i=\frac{\underline{\epsilon}_i+\overline{\epsilon}_i}{2}$, $i=1,\ldots,n$.
i.e., $\epsilon_i$ is the midpoint of interval $[\underline{\epsilon}_i,\overline{\epsilon}_i]$. 
We denote the set of all representative values as $E=\{\epsilon_1,\ldots,\epsilon_{n}\}$.
We define $W_d=\{u+\epsilon \text{ } | \text{ } u \in U,  \epsilon \in E\} \subset W$ as a finite set of applied control inputs. Also, let $\omega: U \rightarrow W_d$ be a random variable, where $\omega(u)=u + \epsilon $ with the probability mass function $p_{\omega}(\omega(u)=u+\epsilon)=\frac{1}{n}$ (follows from Eqn.~(\ref{uniform distribution})). 

 Finally, we define a Quantized System (QS) that approximates the original system as follows: The set of applied control inputs in QS is $W_d$; for a state $q_{k-1}$ and a control input $u_{k} \in U$, QS returns  %the state trajectory at stage $k$:
\begin{equation}
\label{quantized system}
q_k(q_{k-1},\omega(u_k),t)=q_k(q_{k-1},u_k+\epsilon,t)
\end{equation}
with probability $\frac{1}{n}$, where $\epsilon \in E$.

\begin{figure}[htb]
\begin{center}
\includegraphics[width=0.44\textwidth]{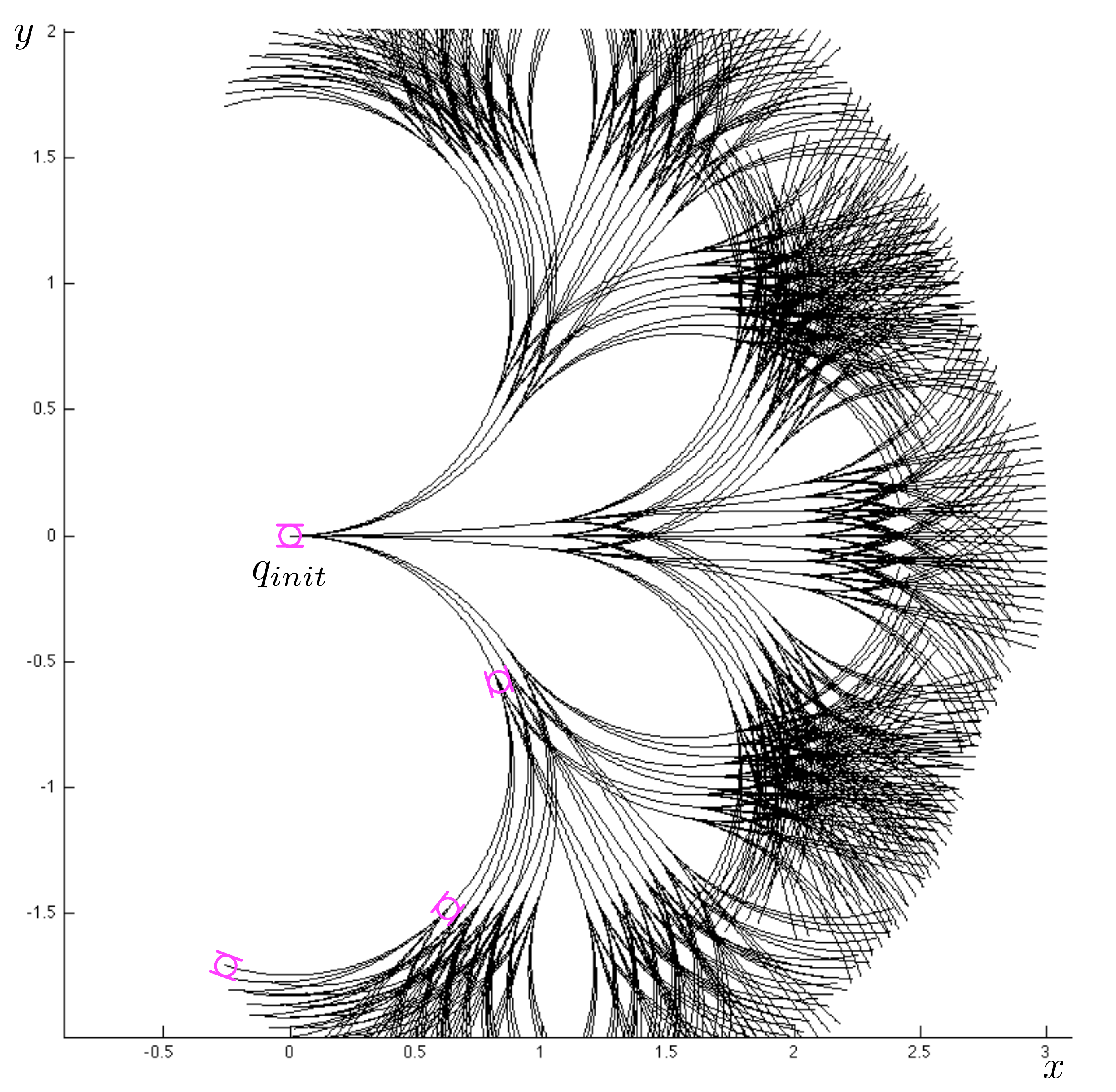}
\end{center}
\caption{The projection of reachability graph $G_3(q_{init})$ on $\mathbb{R}^2$ when $U=\{-\frac{\pi}{3},0,\frac{\pi}{3}\} $ and $E=\{-0.1,0,0.1\}$ with $\Delta t=1.2$. Actual poses of the vehicle are shown in magenta.  }
\label{reachability graph}
\end{figure}

Next, we denote $u_1u_2\ldots u_K$, in which $u_k \in U$ gives a control input at stage $k$, as a finite sequence of control inputs of length $K$. Let $\Sigma_K$ denote the set of all such sequences. For the initial state $q_{init}$ and $\Sigma_K$, we define the reachability graph $G_K(q_{init})$ (see \cite{Lavalle:planning} for a related definition), which encodes the set of all state trajectories originating from $q_{init}$ that can be obtained, with a positive probability, by applying sequences of control inputs from $\Sigma_K$ according to QS given by Eqn.~(\ref{quantized system}) (an example is given in Fig.~\ref{reachability graph}).

\subsection{Position uncertainty and MDP construction} \label{position uncertainty}
As explained before, in order to answer whether some state trajectory satisfies PCTL formula $\phi$ (Eqn.~(\ref{general formula})), it is sufficient to know its projection in $\mathbb{R}^2$. Therefore, we focus only on the position uncertainty. 

\begin{figure}[htb] 
\begin{center}
\includegraphics[width=0.475\textwidth]{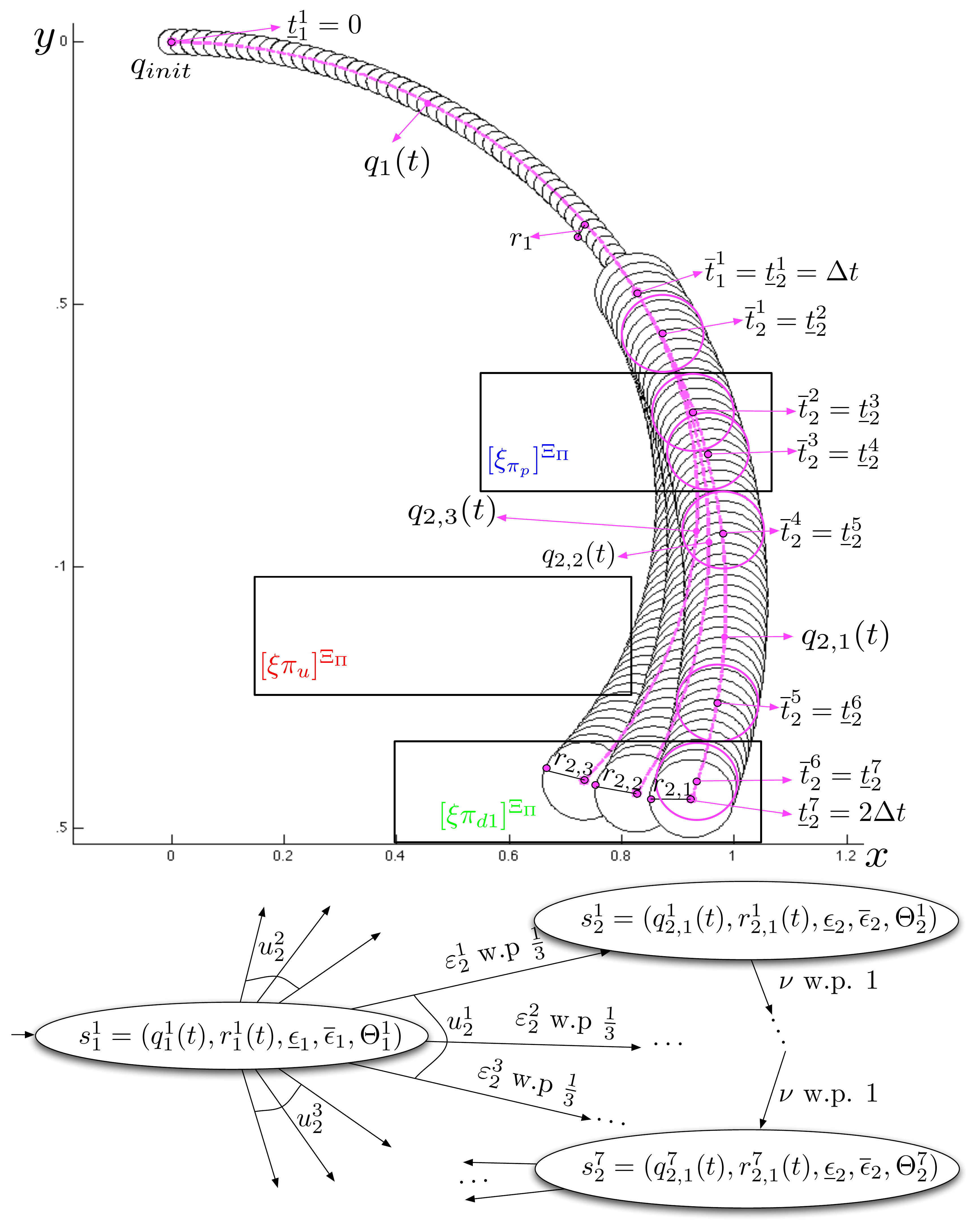}
\end{center}
\caption{Above: An example scenario corresponding to the MDP fragment shown below. $[\xi_ {\pi_u}]^{\Xi_{\Pi}}$, $[\xi_{ \pi_p}]^{\Xi_{\Pi}}$ and $[\xi_{ \pi_{d1}}]^{\Xi_{\Pi}}$ are shown in the figure. Then $[\xi_ { \neg \pi_u}]^{\Xi_{\Pi}}= X \setminus [\xi_ {\pi_u}]$, and similarly for $[\xi_ { \neg \pi_p}]^{\Xi_{\Pi}}$ and $[\xi_ { \neg \pi_{d1}}]^{\Xi_{\Pi}}$ holds. Since along the state trajectory $q_1(t)$ when the uncertainty trajectory is $r_1(t)=r_1$, $t \in [0,\Delta t]$ the set of satisfying propositions does not change, only one state, denoted $s_1^1$, is generated, where $\theta_1 = \{\xi_{\neg \pi_u},\xi_{\neg \pi_p},\xi_{\neg \pi_{d1}}\}$. 
For the state trajectory $q_{2,1}(t)$, when the uncertainty trajectory is $r_{2,1}(t)=r_{2,1}$, $t \in [\Delta t, 2\Delta t]$, the following sequence is generated: $(\Theta_2^1,[\underline{t}_2^1,\overline{t}_2^1]), \ldots, (\Theta_2^7,[\underline{t}_2^7,\overline{t}_2^7])$, where the time interval bounds are shown on the figure and $\Theta_2^1 =\{\xi_{\neg \pi_u},\xi_{\neg \pi_p},\xi_{\neg \pi_{d1}}\}$, $\Theta_2^2 =\{\xi_{\neg \pi_u},\xi_{\neg \pi_{d1}}\}$, $\Theta_2^3 =\{\xi_{\neg \pi_u},\xi_{\pi_p},\xi_{\neg \pi_{d1}}\}$, \ldots, $\Theta_2^6 =\{\xi_{\neg \pi_u},\xi_{\neg \pi_p}\}$ and $\Theta_2^7 =\{\xi_{\neg \pi_u}, \xi_{\neg \pi_p},\xi_{\pi_{d1}}\}$.
 Below: A fragment of the MDP corresponding to the scenario shown above, where $[-\epsilon_{max}, \epsilon_{max}]$ is partitioned into $n=3$ intervals. Action $u_2^1 \in A(s_1^1)$ enables three transitions, each w.p. $\frac{1}{3}$. This corresponds to applied control input being equal to $u_2^1+\epsilon_2^i$ w.p. $\frac{1}{3}$, $\epsilon_2^i \in E$. The elements of $s_2^i$ are: $q_2^i(t)=q_{2,1}(t')$ and $r_2^i(t)=r_{2,1}(t')$, $t' \in [\underline{t}_2^i,\overline{t}_2^i]$, $[\underline{\epsilon}_2,\overline{\epsilon}_2]$  is such that $\epsilon_2^1 \in [\underline{\epsilon}_2,\overline{\epsilon}_2] \in \mathcal{E}$ and $\Theta_2^i$, $i=1,\ldots, 7$. Note that $A(s_2^7)=U$. }
\label{mdp example}
\end{figure}

The position uncertainty of the vehicle when its nominal position is $(x,y) \in \mathbb{R}^2$ is modeled as a disc centered at $(x,y)$ with radius $r \in \mathbb{R}$, where $r$ denotes the uncertainty:
%\begin{equation} \label{position uncertainty}
$D((x,y),r)=\{(x',y') \in \mathbb{R}^2 | ||(x,y), (x',y') || \leq r\},$
%\end{equation}
where $||\cdot||$ denotes the Euclidian distance. The way we model the uncertainty along $q(t) \in G_{K}(q_{init})$ is given in \cite{ProbSafeControlofNoisyDubinsVehicle}. 
Briefly, first, we obtain uncertainty at state $q_k$, denoted $r_k$, by using a worst case scenario assumption: if $u_k + \epsilon_k \in W_d$ is the applied control input for QS,  the corresponding applied control input at stage $k$ for the original system was  $u_k-\underline{\epsilon_k}$ or $u_k+\overline{\epsilon}_k$, where $\epsilon_k \in [\underline{\epsilon_k}, \overline{\epsilon_k}]$.
Then, we define $r : [0,K\Delta t] \rightarrow \mathbb{R}$ as an approximated uncertainty trajectory and we set $r(t)=r_k$, $t \in [(k-1)\Delta t, k\Delta t]$, $k=1,\ldots, K$, i.e., we set the uncertainty along the state trajectory $q_k(t)$ equal to the maximum value of the uncertainty along $q_k(t)$, which is at state $q_k$ (for more details see Fig.~\ref{mdp example}).

A tree-structured  MDP M that models the motion of the vehicle in the environment and the evolution of the position uncertainty is defined as a tuple $(S,s_0, Act, A, P, \Xi_{\Pi}, h)$ where: \\
$\bullet$ $S$ is the finite set of states. The meaning of the state is as follows: $(q(t), r(t), \underline{\epsilon}, \overline{\epsilon}, \Theta) \in S$ means that along the state trajectory $q(t)$, the uncertainty trajectory is $r(t)$; the noise interval is $[\underline{\epsilon}, \overline{\epsilon}] \in \mathcal{E}$; and $\Theta \in 2^{\Xi_{\Pi}}$ is the set of satisfied propositions along the state trajectory $q(t)$ when $r(t)$ is the uncertainty trajectory (see Fig.~\ref{mdp example} for an example). \\
$\bullet$ $s_0=(q_{init},0,0,0, \Theta_{init}) \in S$ is the initial state, where $\Theta_{init} \in 2^{\Xi_{\Pi}}$ is the set of propositions satisfied at $q_{init}$. \\
$\bullet$ $Act = U \cup \nu$ is the set of actions ($\nu$ is a dummy action); \\
$\bullet$ $A: S \rightarrow 2^{Act}$ gives the enabled actions at each state;\\
$\bullet$ $P: S \times Act \times S \rightarrow [0,1]$ is a transition probability function;\\
$\bullet$ $\Xi_{\pi}$ is the set of propositions;\\
$\bullet$ $h : S \rightarrow 2^{\Xi_{\pi}}$ assigns proposition from $\Xi_{\Pi}$ to states $s \in S$ according to the following rule: given  $s = (q(t), r(t), \underline{\epsilon}, \overline{\epsilon}, \Theta) \in S$, $\forall \xi \in \Xi_{\Pi}$, $\xi \in h(s)$ iff $\xi \in \Theta$. 

We generate $S$ and $P$ while building $G_K(q_{init})$ starting from $q_{init}$. Given $q_k(t)=q_k(q_{k-1}, u_k+\epsilon,t) \in G_K(q_{init})$, and the corresponding $r_k(t)$, $t \in [(k-1)\Delta t, k \Delta t]$, $k=1,\ldots,K$, first, we generate a sequence $(\Theta_k^1, [\underline{t}_k^1,\overline{t}_k^1]),\ldots,(\Theta_k^l, [\underline{t}_k^l,\overline{t}_k^l])$, $l \geq 1$, where $\Theta_k^i \in 2^{\Xi_{\Pi}}$ is the set of satisfied propositions along the state trajectory $q_k^i(t)=q_k(t')$, when the corresponding uncertainty trajectory is $r_k^i(t)=r_k(t')$,  for $ t' \in [\underline{t}_k^i,\overline{t}_k^i] \subseteq [(k-1)\Delta t, k \Delta t]$, $i=1,\ldots,l$,  according to the following rules: \\
$\bullet$ Let $\underline{t}_k^1=(k-1)\Delta t$. Then, $D((x_k(\underline{t}_k^1),y_k(\underline{t}_k^1)),r_k(\underline{t}_k^1)) \subseteq [\Theta_k^1]$ and $\overline{t}_k^1= \max_{[\underline{t}_k^1,k\Delta t]}\{t | D((x_k(t),y_k(t),r_k(t) \subseteq [\Theta_k^1]\}$.\\
$\bullet$ If $D((x_k(t),y_k(t),r_k(t)) \subseteq [\Theta_k^{i}]$, $t \in [\underline{t}_k^i,\overline{t}_k^i]$ and $\Theta_k^{i+1} \neq \Theta_k^{i}$, then:
\begin{enumerate}
\item $\exists t \geq \overline{t}_k^i$ s.t.  $D((x_k(t),y_k(t),r_k(t)) \subseteq [\Theta_k^{i+1}]$ and
\item $D((x_k(\tau),y_k(\tau),r_k(\tau)) \nsubseteq [\xi]$, $\forall \tau \in [\overline{t}_k^i,t ]$, $\forall \xi \in \Xi_{\Pi} \setminus (\Theta_k^{i} \cup \Theta_k^{i+1})$.
\item $\underline{t}_k^{i+1}=\overline{t}_k^{i}$ and $\overline{t}_k^{i+1} = \max_{[\underline{t}_k^{i+1},k\Delta t]}\{t | \\
 D((x_k(t),y_k(t),r_k(t) \subseteq [\Theta_k^{i+1}]\}$.
\end{enumerate}
Next, for each $(\Theta_k^i, [\underline{t}_k^i,\overline{t}_k^i])$, $i=1, \ldots, l$, we generate a state of the MDP $s_k^i=(q_k^i(t),r_k^i(t),\underline{\epsilon},\overline{\epsilon},\Theta^k_i)$ such that $q_k^i(t)=q_k(t')$ and $r_k^i(t)=r_k(t')$, $t' \in  [\underline{t}_k^i,\overline{t}_k^i]$ and $\underline{\epsilon}$ and $\overline{\epsilon}$ are such that $\epsilon \in [\underline{\epsilon}, \overline{\epsilon}] \in \mathcal{E}$. Finally, the newly generated state $s_k^i$, $i =1, \ldots l$, $l \geq 1$,  is added to $S$ and the transition probability function is updated, as follows: \\
$\bullet$ If $i < l$, $A(s_k^i)=\nu$ and $P(s_k^i,\nu,s_k^{i+1})=1$, and otherwise, i.e., if $i=l$, $A(s_k^l)=U$ and $\forall u_{k+1} \in U$, $P(s_k^l,u_{k+1},s_{k+1}^1)=\frac{1}{n}$. 

The former follows from the fact that $k\Delta$ is not reached and control input for the next stage needs not to be chosen. Under dummy action $\nu$, with probability $1$, the system makes a transition to the next state in the sequence satisfying a different set of propositions. 
The latter follows from the fact that $k\Delta t$ is reached and the control input for the next stage needs to be chosen. Given a control input $u_{k+1} \in U$ the applied control input will be $u_k+\epsilon \in W_d$, $\epsilon \in E$, with probability $\frac{1}{n}$, and given a new state trajectory $q_{k+1}(q_k,u_{k+1}+\epsilon,t)$ (Eqn.~(\ref{quantized system})) the first corresponding state will be  $s_{k+1}^1$ (see Fig.~\ref{mdp example}).\\
$\bullet$ If the termination time is reached, we set $A(s_k^i)=\nu$ and $P(s_k^i,\varphi,s_k^i)=1$. Such state is called a $leaf$ state.

\begin{prop} \label{valid mdp}
The model $M$ defined above is a valid tree-structured MDP, i.e., it satisfies the Markov property, $P$ is a valid transition probability function and each state has exactly one incoming transition. 
\end{prop}
{\bf{Proof:}} The proof follows from the construction of the MDP. Given a current state $s \in S$ and an action $a \in A(s)$, the conditional probability distribution of future states depends only on the current state $s$, not on the sequences of events that proceed it (see the rules stated above). Thus, Markov property holds. In addition, since $\sum_{\epsilon \in \mathcal{E}} p_{\omega}(\omega(u)=u+\epsilon)=1$, it follows that $P$ is a valid transition probability function. Finally, the fact that M is a tree-structured MDP follows from the following: for each $q_k(t) \in G_K({q_{init}})$,  a unique sequence of states $s_k^i$, $i=1,\ldots,l$, $l \geq 1$ is generated. Each state in that sequence has exactly one incoming transition. Thus, according to Def.~\ref{tree structured map}, $M$ is a tree-structured MDP. $\blacksquare$

\section{PCTL Control Policy Generation} \label{pctl control policy generation}

\subsection{Control policy for the initial PCTL formula}  \label{offline control policy}
The proposed PCTL control synthesis is an adaptation of the approach from \cite{LWAB10}. Specifically, we exploit the tree-like structure of $M$ and develop an efficient algorithm for generating a control policy for $M$ that maximizes the probability of satisfying a PCTL formula ${\bm {\phi}}$ (Eqn.~(\ref{general formula})). 

Given a tree-structured MDP $M=(S,s_0,A,Act,P,\Xi_{\Pi}, h)$ and a PCTL formula ${\bm {\phi}} : = \mathcal{P}_{max=?} [\mathcal{P}_{\geq p_1}[\varphi_1 \mathcal{U} ( \psi_1 \wedge \mathcal{P}_{\geq p_2}[\varphi_2 \mathcal{U} ( \psi_2 \wedge\ldots \wedge \mathcal{P}_{\geq p_f}[\varphi_f \mathcal{U} \psi_f])])]]$, we are interested in obtaining the control policy $\mu_{\bm{\phi}}$ that maximizes the probability of satisfying ${\bm{\phi}}$, as well as the corresponding probability value, denoted $V_{\bm{\phi}}$, where $V_{\bm{\phi}}: S \rightarrow [0,1]$. Specifically, for $s \in S$, $\mu_{\bm{\phi}}(s) \in A(s)$ is the action to be applied at $s$ and $V_{\bm{\phi}}(s)$ is the probability of satisfying ${\bm{\phi}}$ at $s$ under control policy $\mu_{\bm{\phi}}$. 
To solve this problem we propose the following approach:

{\emph{Step 1:}} Solve ${\bm{\phi}_f}:=\mathcal{P}_{\geq p_f}[\varphi_f \mathcal{U} \psi_f]$, i.e., find the set of initial states $S_{{\bm{\phi}_f}}$ from which ${\bm{\phi}_f}$ is satisfied with probably greater than or equal to $p_f$ and determine the corresponding control policy $\mu_{\bm{\phi}_f}$. To solve this problem, first, let ${\bm{\phi}_f'}:=\mathcal{P}_{\max=?}[\varphi_f \mathcal{U} \psi_f]$, and compute the maximizing probabilities $V_{\bm{\phi}'_f}$. This can be done by dividing $S$ into three subsets $S_{\bm{\phi}_f'}^{yes}$ (states satisfying ${\bm{\phi}_f'}$ with probability $1$), $S_{\bm{\phi}_f'}^{no}$ (states satisfying ${\bm{\phi}_f'}$ with probability $0$), and $S_{\bm{\phi}_f'}^?$ (the remaining states): 
\begin{equation*}
{\small{
\begin{split}
S_{\bm{\phi}_f'}^{yes} &=\text{Sat}(\psi_f),\\
S_{\bm{\phi}_f'}^{no} &= S \setminus ( \text{Sat}(\varphi_f )\cup \text{Sat}(\psi_f)),\\
S_{\bm{\phi}_f'}^{?} &= S \setminus (S_{\bm{\phi}_f'}^{yes} \cup S_{\bm{\phi}_f'}^{no}),
\end{split}
}}
\end{equation*}
where $\text{Sat}(\psi_f)$ and $\text{Sat}(\varphi_f)$ are the set of states satisfying $\psi_f$ and $\varphi_f$, respectively.
The computation of maximizing probabilities for the states in $S$ can be obtained as a unique solution of the following system: 
\begin{equation} \label{maximizing probabilities}
{\small{
V_{\bm{\phi}_f'}(s) =	\left\{ \begin{array}{ccl} 1 & \mbox{if} & s \in S_{\bm{\phi}_f'}^{yes} \\ 0 & \mbox{if} & s \in S_{\bm{\phi}_f'}^{no} \\ \text{max}_{a \in A(s)}\{\sum_{s' \in S}P(s,a,s')V_{\bm{\phi}_f'}(s')\}  & \mbox{if} & s \in S_{\bm{\phi}_f'}^{?}
\end{array}\right.
}}
\end{equation}
and the control policy at each state is equal to the action that gives rise to this optimal solution, i.e., $\forall s \in S$, $\mu_{\bm{\phi}_f'}(s)=\text{argmax}_{a \in A(s)}\{\sum_{s' \in S}P(s,a,s')V_{\bm{\phi}_f'}(s')\}$.

In general (i.e., for a non tree-structured MDPs containing cycles), solving Eqn.~(\ref{maximizing probabilities}) requires solving a linear programming problem (\cite{baier:principles, LWAB10}). For a tree-structured MDPs the solution can be obtained in a simple fashion: from each leaf state of the MDP, move backwards, by visiting parent states until $s_0$ is reached; at each state in $S_{\bm{\phi}_f'}^{?}$ perform maximization from Eqn.~(\ref{maximizing probabilities}). The fact that $M$ contains no cycles is sufficient to see that the procedure stated above will result in maximizing probabilities. 

The state formula ${\bm{\phi}_f}$ requires to reach a state in $\text{Sat}(\psi_f)$ by going through states in $\text{Sat}(\varphi_f)$ with probability greater than or equal to $p_f$. Thus, $\forall s \in S$ s.t. $V_{\bm{\phi}_f'}(s) < p_f$ we set $V_{\bm{\phi}_f}(s)=0$, and otherwise, i.e., $\forall s \in S$ s.t. $V_{\bm{\phi}_f'}(s) \geq p_f$ we set $V_{\bm{\phi}_f}(s)=V_{\bm{\phi}_f'}(s)$. Finally, $\forall s \in S$,  $\mu_{\bm{\phi}_f}(s)=\mu_{\bm{\phi}_f'}(s)$ and the set of initial states is $S_{{\bm{\phi}_f}}=\{s \in S | V_{\bm{\phi}_f} (s)> 0\}.$ 

{\emph{Step 2:}} Solve ${\bm{\phi_{f-1}}}:=\mathcal{P}_{\geq p_{f-1}}[\varphi_f \mathcal{U} (\psi_{f-1} \wedge {\bm{\phi}_f})]$, i.e., find the set of initial states $S_{{\bm{\phi_{f-1}}}}$ from which ${\bm{\phi_{f-1}}}$ is satisfied with probability greater than or equal to $p_{f-1}$. To solve this problem, again, begin by solving ${\bm{\phi_{f-1}'}}:=\mathcal{P}_{\max=?}[\varphi_{f-1} \mathcal{U} (\psi_{f-1} \wedge {\bm{\phi}_f}) ]$. Start by dividing $S$ into three subsets: 
\begin{equation*}
{\small{
\begin{split}
S_{\bm{\phi}_{f-1}'}^{yes} &=\text{Sat}(\psi_{f-1}) \cap S_{{\bm{\phi}_f}},\\
S_{\bm{\phi}_{f-1}'}^{no} &= S \setminus ( \text{Sat}(\varphi_{f-1} )\cup S_{\bm{\phi}_{f-1}'}^{yes})),\\
S_{\bm{\phi}_{f-1}'}^{?} &= S \setminus (S_{\bm{\phi}_{f-1}'}^{yes} \cup S_{\bm{\phi}_{f-1}'}^{no}),
\end{split}
}}
\end{equation*}
Note that,  $S_{\bm{\phi}_{f-1}'}^{yes}$ is the set of states satisfying $\psi_{f-1}$ intersected with $S_{{\bm{\phi}_f}}$.
Next, perform the same procedure as in {\emph{Step 1}} for obtaining $V_{\bm{\phi}_{f-1}}$, $\mu_{\bm{\phi}_{f-1}}$ and $S_{{\bm{\phi_{f-1}}}}$.

{\emph{Step 3:}} Repeat {\emph{Step 2}} for ${\bm{\phi_{f-2}}}, {\bm{\phi_{f-3}}}, \ldots, {\bm{\phi_{1}}}$, i.e., until $V_{\bm{\phi}_{1}}$, $\mu_{\bm{\phi}_{1}}$ and $S_{{\bm{\phi_{1}}}}$ are obtained where ${\bm{\phi_{1}}}:=\mathcal{P}_{\geq p_{1}}[\varphi_1 \mathcal{U} (\psi_1 \wedge {\bm{\phi}_2})]$.

%It should be noted that, by the nature of the PCTL formulas, the execution of optimal control policy $\mu_{\bm{\phi}}$ guarantees the satisfaction of ${\bm{\phi}}$ only if the  system reaches a state in $\text{Sat}(\psi_1)$ by going through the states in $\text{Sat}(\varphi_1)$ with probability greater than or equal to $p_1$, 
%%and then reaches a state in $\text{Sat}(\psi_{2})$ by going through the states in $\text{Sat}(\varphi_{2})$ with probability greater than or equal to $p_{2}$, 
%and then $\cdots$ reaches a state in $\text{Sat}(\psi_f)$ by going through the states in $\text{Sat}(\varphi_f)$ with probability greater than or equal to $p_f$. 

By the nature of the PCTL formulas, to ensure the execution of all specified tasks in ${\bm{\phi}}$, we construct a history dependent control policy of the following form: \\
{\emph{ $\mu_{\bm{\phi}}:$ Apply policy $\mu_{\bm{\phi}_{1}}$ until a state in  $S_{\bm{\phi}_{1}'}^{yes}$ is reached. Then, apply policy $\mu_{\bm{\phi}_{2}}$ until a state in  $S_{\bm{\phi}_{2}'}^{yes}$ is reached. 
$\cdots$  
Finally, apply $\mu_{\bm{\phi}_{f}}$ until a state in $S_{\bm{\phi}_{f}'}^{yes}$ is reached. 
}}

For the same reason as stated above, $V_{\bm{\phi}}(s_0)$, the maximum probability of satisfying ${\bm{\phi}}$, can not be found directly because it is not known which state in $S_{\bm{\phi}_{1}'}^{yes}, S_{\bm{\phi}_{2}'}^{yes}, \ldots, S_{\bm{\phi}_{f}'}^{yes}$ will be reached first. However, since the probability of satisfying ${\bm{\phi}_{i}}$ from each state in $S_{\bm{\phi}_{i-1}'}^{yes}$ is available, a bound on the probability of satisfying ${\bm{\phi}}$ can be defined. The lower and upper bounds are $V_{\bm{\phi}_1}(s_0) \cdot V_{\bm{\phi}_2}^{min} \cdot \ldots \cdot V_{\bm{\phi}_f}^{min}$ and $V_{\bm{\phi}_1}(s_0) \cdot V_{\bm{\phi}_2}^{max} \cdot \ldots \cdot V_{\bm{\phi}_f}^{max}$, where $V_{\bm{\phi}_i}^{min}$ and $V_{\bm{\phi}_i}^{max}$ denote the minimum and maximum probability of satisfying ${\bm{\phi}_i}$ from $S_{\bm{\phi}_{i-1}'}^{yes}$.

In \cite{ProbSafeControlofNoisyDubinsVehicle} we show that a sequence of measured intervals corresponds to a unique state on the MDP. Thus, the desired vehicle control strategy $\Gamma_{\bm {\phi}}$ returns the control input for the next stage by mapping the sequence to the state of the MDP; the control input corresponds to the optimal action, under $\mu_{\bm{\phi}}$, at that state. %The main result from \cite{ProbSafeControlofNoisyDubinsVehicle}  naturally extends to the following result:
%\begin{thm}
%\label{theorem}
%The probability that the system given by Eqn. (\ref{dubins kinematics}), under the control strategy $\Gamma_{\bm {\phi}}$, generates a trajectory that satisfies a PCTL formula ${\bm {\phi}}$ (Eqn.~(\ref{general formula})) is bounded from below by $V_{\bm {\phi}}(s_0)$, where $V_{\bm {\phi}}(s_0)$, the probability of satisfying ${\bm {\phi}}$ on the MDP under the control policy $\mu_{\bm{\phi}}$, is bounded from below by $V_{\bm{\phi}_1}(s_0) \cdot V_{\bm{\phi}_2}^{min} \cdot \ldots \cdot V_{\bm{\phi}_f}^{min}$.
%\end{thm}

\subsection{Control policy  for the updated PCTL formula} \label{online control policy}
Next, assume that at the end of stage $k$, for some $k=0,\ldots,K-1$,  ${\bm {\phi}}$ is updated into ${\bm {\phi}}^+$. As noted in the previous subsection, given a sequence of measured intervals, we can follow vehicle's progress on $M$. We denote the current state as $s_{C} \in S$ (if it is at the initial state, then $s_C=s_0$).  We develop an efficient algorithm for obtaining $\mu_{\bm {\phi}^+}$, % that maximizes the probability of satisfying ${\bm {\phi}^+}$ 
and $V_{\bm {\phi}^+}$, that reuses $\mu_{\bm {\phi}}$ and $V_{\bm {\phi}}$, and exploits the structure of formulas given by Eqn.~(\ref{general formula}) and the fact that $M$ is a tree-structured MDP. 

%First, we formally define the ways in which ${\bm {\phi}}$ can be updated. Since $\forall j \in \{1,\ldots, f\}$, $\psi_j$ and $\varphi_j$ are in CNF and DNF, respectively, they can be expressed as $$\psi_j = (\bigvee_{\xi \in \Xi_{\psi,j}^1}\xi) \wedge \ldots \wedge (\bigvee_{\xi \in \Xi_{\psi,j}^{n_{j}}}\xi)$$ and $$\varphi_j = (\bigwedge_{\xi \in \Xi_{\varphi,j}^1}\xi) \vee \ldots \vee (\bigwedge_{\xi \in \Xi_{\varphi,j}^{m_j}}\xi),$$ where $\forall_{j=1,\ldots,f}n_j,m_j \in \mathbb{Z}^+$, $\forall_{n=1,\ldots,n_j} \Xi_{\psi,j}^n \subseteq \Xi_{\Pi}$, $\forall_{m=1,\ldots,n_j} \Xi_{\varphi,j}^n \subseteq \Xi_{\Pi}$. 

First, we formally define what it means for ${\bm {\phi}}$ to be satisfied up to $i$, $0 \leq i \leq f$. Note that, if under the execution of $\mu_{\bm {\phi}}$, $S_{\bm{\phi}_{i}'}^{yes}$ is reached, it is guaranteed that $\mathcal{P}_{\geq p_1}[\varphi_1 \mathcal{U} ( \psi_1 \wedge \ldots \wedge \mathcal{P}_{\geq p_i}[\varphi_i \mathcal{U} \psi_i])]$ part of ${\bm {\phi}}$  is satisfied. Thus, ${\bm {\phi}}$ is satisfied up to $i$, where $i=\max_{j \in \{0,\ldots,f\}}\{j | S_{\bm{\phi}_{j}'}^{yes} \text{ is reached}\}$, $S_{\bm{\phi}_{0}'}^{yes}=s_0$. 

Next, since $\forall j \in \{1,\ldots, f\}$,   $\varphi_j$ and $\psi$ are in CNF and DNF, respectively, they can be expressed as $\varphi_j = \varphi_j^1 \wedge \ldots \wedge \varphi_j^{m_j}$ and $\psi_j = \psi_j^1 \vee \ldots \vee \psi_j^{n_j}$ where $m_j, n_j \in \mathbb{Z^+}$ and $\forall_{m=1,\ldots,m_j} \varphi_j^m$ is a disjunction clause (disjunction of propositions from $\Xi_{\Pi}$) and $\forall_{n=1,\ldots,n_j} \psi_j^n$ is a conjunction clause (conjunction of propositions from $\Xi_{\Pi}$). We are ready to formulate the specification update rules:

{{\emph{Specification update rules:}} Given ${\bm {\phi}}$ satisfied up to $i$, $0 \leq i \leq f$, the updated formula ${\bm {\phi}}^+$ is obtained from ${\bm {\phi}}$ by removing $\mathcal{P}_{\geq p_1}[\varphi_1 \mathcal{U} ( \psi_1 \wedge \ldots \wedge \mathcal{P}_{\geq p_i}[\varphi_i \mathcal{U} \psi_i])]$ from ${\bm {\phi}}$, and then by updating $\psi_j$, $\varphi_j$, or $p_j$ for $j \in \{i,\ldots,f\}$:\\
%\begin{enumerate}
%\item  $\psi_j^+=\psi_j^1 \vee \ldots \vee \psi_j^{n_j+1}$; or
%\item  $\psi_j^+=\psi_j^1 \vee \ldots \vee \psi_j^{n_j-1}$, if $n_j \geq 1$; or
%\item $\varphi_j^+=\varphi_j^1 \wedge \ldots \wedge \varphi_j^{m_j-1}$, if $m_j \geq 1$; or
%\item $\varphi_j^+=\varphi_j^1 \wedge \ldots \wedge \varphi_j^{m_j+1}$; or
%\item $p_j^+ \in [0,1]$ s.t. $p_j^+ > p_j$; or
%\item $p_j^+ \in [0,1]$ s.t. $p_j^+ < p_j$.
%\end{enumerate}
1) $\psi_j^+=\psi_j^1 \vee \ldots \vee \psi_j^{n_j+1}$; or\\
2) $\psi_j^+=\psi_j^1 \vee \ldots \vee \psi_j^{n_j-1}$, if $n_j \geq 1$; or\\
3) $\varphi_j^+=\varphi_j^1 \wedge \ldots \wedge \varphi_j^{m_j-1}$, if $m_j \geq 1$; or\\
4) $\varphi_j^+=\varphi_j^1 \wedge \ldots \wedge \varphi_j^{m_j+1}$; or\\
5) $p_j^+ \in [0,1]$ s.t. $p_j^+ < p_j$; or\\
6) $p_j^+ \in [0,1]$ s.t. $p_j^+ > p_j$;
 where $\psi_j^{n_j+1}$ and $\varphi_j^{m_j-1}$ are conjunction and disjunction clauses from $\Xi_{\Pi}$, respectively.

First, note that since $M$ is a tree-structured MDP, $\mu_{\bm {\phi}^+}$ needs to be defined only for the states reachable from current state $s_C \in S$. Thus, we construct a new tree-structured MDP $M^+ \subseteq M$, for which $s_C$ is the initial state, by eliminating the states that are not reachable form $s_C$. For a tree-structured MDP this is a straightforward process. 
Next, we use the approach presented in Sec.~\ref{offline control policy} and show that we can partially reuse  $\mu_{\bm {\phi}}$ and $V_{\bm {\phi}}$ when solving the problem. Additionally, we show that for updates 1), 3), and 5), $\forall s \in S^+$, $V_{\bm {\phi}^+}(s) \geq V_{\bm {\phi}}(s) $, and that for updates 2), 4), and 6), $\forall s \in S^+$, $V_{\bm {\phi}^+}(s) \leq V_{\bm {\phi}}(s)$, where $S^+$ is the set of states of $M^+$. %To be able to do this comparison, for the rest of the paper we assume that $\mathcal{P}_{\geq p_1}[\varphi_1 \mathcal{U} ( \psi_1 \wedge \ldots \wedge \mathcal{P}_{\geq p_i}[\varphi_i \mathcal{U} \psi_i])]$ is removed from ${\bm {\phi}}$ as well. 

{{\emph{Update 1}}: Adding a conjunction clause $\psi_j^{n_j+1}$ to  $\psi_j$, resulting in $\psi_j^+=\psi_j \vee \psi_j^{n_j+1}$. %We solve ${\bm {\phi}}^+$  on $M^+$ by using approach presented in  in Sec.~\ref{offline control policy}. 
Since for $k \in \{j+1, \ldots, f\}$, ${\bm {\phi}_{k}^+}={\bm {\phi}_{k}}$, it follows that $\mu_{\bm {\phi}_{k}^+}=\mu_{\bm {\phi}_{k}}$, $V_{\bm {\phi}_{k}^+}=V_{\bm {\phi}_{k}}$, and $S_{\bm {\phi}_{k}^+}=S_{\bm {\phi}_{k}}$ (this holds for all other updates as well). 
When solving ${\bm {\phi}_{j}^+}:=\mathcal{P}_{\geq p_j}[\varphi_{j} \mathcal{U} ((\psi_{j} \vee \psi_j^{n_j+1}) \wedge {\bm{\phi}_{j+1}}) ]$, i.e., in particular ${\bm {\phi}_{j}^{+'}}:=\mathcal{P}_{\max=?}[\varphi_{j} \mathcal{U} ((\psi_{j} \vee \psi_j^{n_j+1}) \wedge {\bm{\phi}_{j+1}}) ]$ note that: 
\begin{equation*}
{\small{
\begin{split}
S_{\bm {\phi}_{j}^{+'}}^{yes} &=(\text{Sat}(\psi_{j}) \cap S_{{\bm{\phi}_{j+1}}}) \cup (\text{Sat}(\psi_{j}^{n_j+1}) \cap S_{{\bm{\phi}_{j+1}}}),\\
S_{\bm {\phi}_{j}^{+'}}^{no} &= S \setminus ( \text{Sat}(\varphi_{j} )\cup S_{\bm {\phi}_{j}^{+'}}^{yes})),\\
S_{\bm {\phi}_{j}^{+'}}^{?} &= S \setminus (S_{\bm{\phi}_{j}'}^{yes} \cup S_{\bm{\phi}_{j}'}^{no}).
\end{split}
}}
\end{equation*}
By using Eqn.~(\ref{maximizing probabilities}) we obtain $\mu_{\bm {\phi}_{j}^{+'}}$ and $V_{\bm {\phi}_{j}^{+'}}$, and then
$\mu_{\bm {\phi}_{j}^{+}}$, $V_{\bm {\phi}_{j}^{+}}$ and $S_{\bm {\phi}_{j}^{+}}$ as described in Sec.~\ref{offline control policy}. From the fact that  $S_{\bm {\phi}_{j}^{+'}}^{yes} \supseteq S_{\bm {\phi}_{j}{'}}^{yes}$ it follows that $\forall s \in S^+$, $V_{\bm {\phi}_{j}^{+'}}(s) \geq V_{\bm {\phi}_{j}{'}}(s)$, and thus $S_{\bm {\phi}_{j}^{+}} \supseteq S_{\bm {\phi}_{j}} $ and $V_{\bm {\phi}_{j}^{+}}(s) \geq V_{\bm {\phi}_{j}}(s)$. This property holds all the way down until $\mu_{\bm {\phi}_{i+1}^{+}}$ and $V_{\bm {\phi}_{i+1}^{+}}$ are obtained. Therefore, $\forall s \in S^+$, $V_{\bm {\phi}^{+}}(s) \geq V_{\bm {\phi}}(s)$. 

{{\emph{Update 2}}: Removing a conjunction clause $\psi_j^{n_j}$ from $\psi_j$, resulting in $\psi_j^+=\psi_j \vee \ldots \vee \psi_j^{n_j-1}$. We follow the approach from {\emph{Update 1}}, with the final result being: $\forall s \in S^+$, $V_{\bm {\phi}^{+}}(s) \leq V_{\bm {\phi}}(s)$, which follows from the fact that $S_{\bm {\phi}_{j}^{+'}}^{yes} \subseteq S_{\bm {\phi}_{j}{'}}^{yes}$.% and then the same reasoning as above is used. 

{{\emph{Update 3}}: Removing a disjunction clause $\varphi_j^{m_j}$ from $\varphi_j$, resulting in $\varphi_j^+=\varphi_j^1 \wedge \ldots \wedge \varphi_j^{m_j-1}$. We follow the approach from {\emph{Update 1}}, with the final result being: 
$\forall s \in S^+$, $V_{\bm {\phi}^{+}}(s) \geq V_{\bm {\phi}}(s)$, which follows from the fact that $S_{\bm {\phi}_{j}^{+'}}^{no} =  S \setminus ( \text{Sat}(\varphi_{j}^{+} )\cup S_{\bm {\phi}_{j}^{+'}}^{yes})) \subseteq S_{\bm {\phi}_{j}{'}}^{no} = S \setminus ( \text{Sat}(\varphi_{j} )\cup S_{\bm {\phi}_{j}^{+'}}^{yes}))$, since $\text{Sat}(\varphi_{j}^{+}) \supseteq \text{Sat}(\varphi_{j} )$.%, and then the same reasoning as before is used. 

{{\emph{Update 4}}: Adding a disjunction clause $\varphi_j^{m_j+1}$ to $\varphi_j$, resulting in $\varphi_j^+=\varphi_j \wedge\varphi_j^{m_j+1}$.
We follow the approach from {\emph{Update 1}}, with the final result being: $\forall s \in S^+$, $V_{\bm {\phi}^{+}}(s) \leq V_{\bm {\phi}}(s)$, which follows from the fact that $S_{\bm {\phi}_{j}^{+'}}^{no} =  S \setminus ( \text{Sat}(\varphi_{j}^{+} )\cup S_{\bm {\phi}_{j}^{+'}}^{yes})) \supseteq S_{\bm {\phi}_{j}{'}}^{no} = S \setminus ( \text{Sat}(\varphi_{j} )\cup S_{\bm {\phi}_{j}^{+'}}^{yes}))$, since $\text{Sat}(\varphi_{j}^{+}) \subseteq \text{Sat}(\varphi_{j} )$.%, and then the same reasoning as before is used. 

{{\emph{Update 5}}: Decreasing $p_j$ such that $p_j^+ \leq p_j$, $p_j^+ \in [0,1]$.
We follow the approach from {\emph{Update 1}}, with the final result being: $\forall s \in S^+$, $V_{\bm {\phi}^{+}}(s) \geq V_{\bm {\phi}}(s)$, which follows from the fact that $S_{\bm {\phi}_{j}^{+}} \supseteq S_{\bm {\phi}_{j}}$, and then 
$V_{\bm {\phi}_{j-1}^{+}}(s) \geq V_{\bm {\phi}_{j-1}}(s)$. %From here the same reasoning as before is used. 
The fact that $S_{\bm {\phi}_{j}^{+}} \supseteq S_{\bm {\phi}_{j}}$ follows from: $S_{{\bm{\phi}_j^+}}=\{s \in S^+ | V_{\bm{\phi}_j^{+'}} (s)> p_j^+\}$ and $S_{{\bm{\phi}_j}}=\{s \in S^+ | V_{\bm{\phi}_j'} (s)> p_j\}$ where $\forall s \in S^+$, $V_{\bm {\phi}_{j}^{+'}}(s) = V_{\bm {\phi}_{j}'}(s)$, and  $p_j^+ \leq p_j$.

{{\emph{Update 6}}: Increasing $p_j$ such that $p_j^+ \geq p_j$, $p_j^+ \in [0,1]$.
We follow the approach from {\emph{Update 5}}, with the final result being: $\forall s \in S^+$, $V_{\bm {\phi}^{+}}(s) \leq V_{\bm {\phi}}(s)$, which follows from the fact that $S_{\bm {\phi}_{j}^{+}} \subseteq S_{\bm {\phi}_{j}}$ and then 
$V_{\bm {\phi}_{j-1}^{+}}(s) \leq V_{\bm {\phi}_{j-1}}(s)$ (see {{\emph{Update 5)}}). %From here the same reasoning as in {\emph{Update e)}} is used. 

For the reasons stated in the previous subsection $\mu_{\bm {\phi}^+}$  has also a history dependent form and we can find the lower and upper bounds of $V_{\bm {\phi}^+}(s_C)$.
%For the reasons stated in the previous subsection $\mu_{\bm {\phi}^+}$ is of the following form: {\emph{ $\mu_{\bm {\phi}^+}:$ Apply policy $\mu_{\bm{\phi}^+_{i+1}}$ until a state in  $S_{\bm{\phi}^{+'}_{i+1}}^{yes}$ is reached. $\cdots$ Finally, apply policy $\mu_{\bm{\phi}^+_{f}}$ until a state in  $S_{\bm{\phi}^{+'}_{f}}^{yes}$ is reached.}} Similarly, $V_{\bm{\phi}^+_{i+1}}(s_C) \cdot V_{\bm{\phi}_{i+2}}^{min} \cdot \ldots \cdot V_{\bm{\phi}_f}^{min}$ and $V_{\bm{\phi}^+_{i+1}}(s_C) \cdot V_{\bm{\phi}_{i+2}}^{max} \cdot \ldots \cdot V_{\bm{\phi}_f}^{max}$ are the lower and upper bounds of $V_{\bm {\phi}^+}(s_C)$.  Once $\mu_{\bm {\phi}^+}$ is obtained it is mapped to the updated vehicle control strategy, denoted $\Gamma_{\bm {\phi}^+}$.% through the procedure described in Sec.~\ref{control strategy}. 

\section{Case study} \label{case study}
We considered the system given by Eqn. (\ref{dubins kinematics}) and we used the following numerical values: $1/\rho=\pi/3$, $\Delta t=1.2$, $K=9$, and $\epsilon_{max}=0.06$ with $n=3$, i.e., $\Delta \epsilon=0.04$. Thus, the maximum actuator noise was approximately $6\%$ of the maximum control input. Three cases are shown in Fig.~\ref{simulation results}.

{\emph{Offline phase:}}
Cases $A$ and $B$ correspond to the offline phase. Initially, the motion specification was as given in Example 1, and the corresponding PCTL formula was $\phi$ (Eqn.~\ref{sample formula 1}). The lower bound on the probability of satisfying $\phi$ on the corresponding MDP was $0.68$. For case $A$ we assumed that the supervisor was satisfied with the satisfaction probability and the vehicle was deployed under the obtained vehicle control strategy. Case $B$ corresponds to the case when the user is not satisfied with a satisfaction probability of 0.68. Then, the system generated a set of specification relaxations, based on the specification update rules from Sec.~\ref{online control policy}, that guaranteed an increase in the satisfaction probability.  We assumed that the supervisor agreed with the specification which ``allowed the vehicle to go through a {\ttfamily{test1}} region before entering a {\ttfamily{pick-up}} region'' (corresponds to {\emph{Update 3}}),  with the corresponding satisfaction probability being $0.85$ ({\emph{Update 3}} increases the satisfaction probability). %Then, the vehicle was deployed under the updated vehicle control strategy. 
    
%
%In order to verif y Theorem 1, we simulated the system under the obtained vehicle control strategy. The simulation based satisfaction probability (number of satisfying trajectories over the number of violating trajectories) was $0.74$ (for 5000 generated trajectories). The result supports Theorem 1, since the simulation based probability is bounded from below by the theoretical probability. 

{\emph{Online phase:}}
Case $C$ corresponds to the online phase. The vehicle was deployed under the initial vehicle control strategy from case $A$ and at $5\Delta t$ the {\ttfamily{drop-off2}} regions became  unavailable for the drop off, and thus the updated specification ``allowed the vehicle to drop off the load only in the {\ttfamily{drop-off1}} regions'' (corresponds to {\emph{Update 2}}). The updated satisfaction probability, returned by the the control synthesis part, was $0.63$ ({\emph{Update 2}} reduces the satisfaction probability). Assuming that the supervisor was satisfied with the updated satisfaction probability the vehicle continued the deployment, now under the updated vehicle control strategy.  

To verify the fact that the result from \cite{ProbSafeControlofNoisyDubinsVehicle} (i.e., the theoretical satisfaction probability is lower bound for the actual satisfaction probability (see Sec.~\ref{problem formulation and approach})), naturally extends to this work, we simulated the original system under the obtained vehicle control strategies. The simulation based satisfaction probabilities (number of satisfying over the number of generated trajectories) for cases $A$, $B$, and $C$, were $0.74$, $0.92$ and $0.72$, respectively. Since the simulation based probabilities are bounded from below by the satisfaction probabilities obtained on the MDP, the result holds.

The constructed MDP had approximately 45000 states. The Matlab code used to construct the MDP ran for 8 min and 52 sec on a computer with a 2.5GHz dual processor. The control synthesis algorithm for case $A$ (initial PCTL control policy generation) ran for  23 sec. For cases $B$ and $C$ (updated PCTL control policy generation) the control synthesis algorithm ran for $11$ and $6$ seconds, respectively. In cases $B$ and $C$ the running time improved by reusing the initial solution from case $A$. There was an additional improvement in case $C$ since the vehicle was moving prior to the update and the updated solution was obtained on the reduced MDP.

\begin{figure}[htb] 
\begin{center}
\includegraphics[width=0.454 \textwidth]{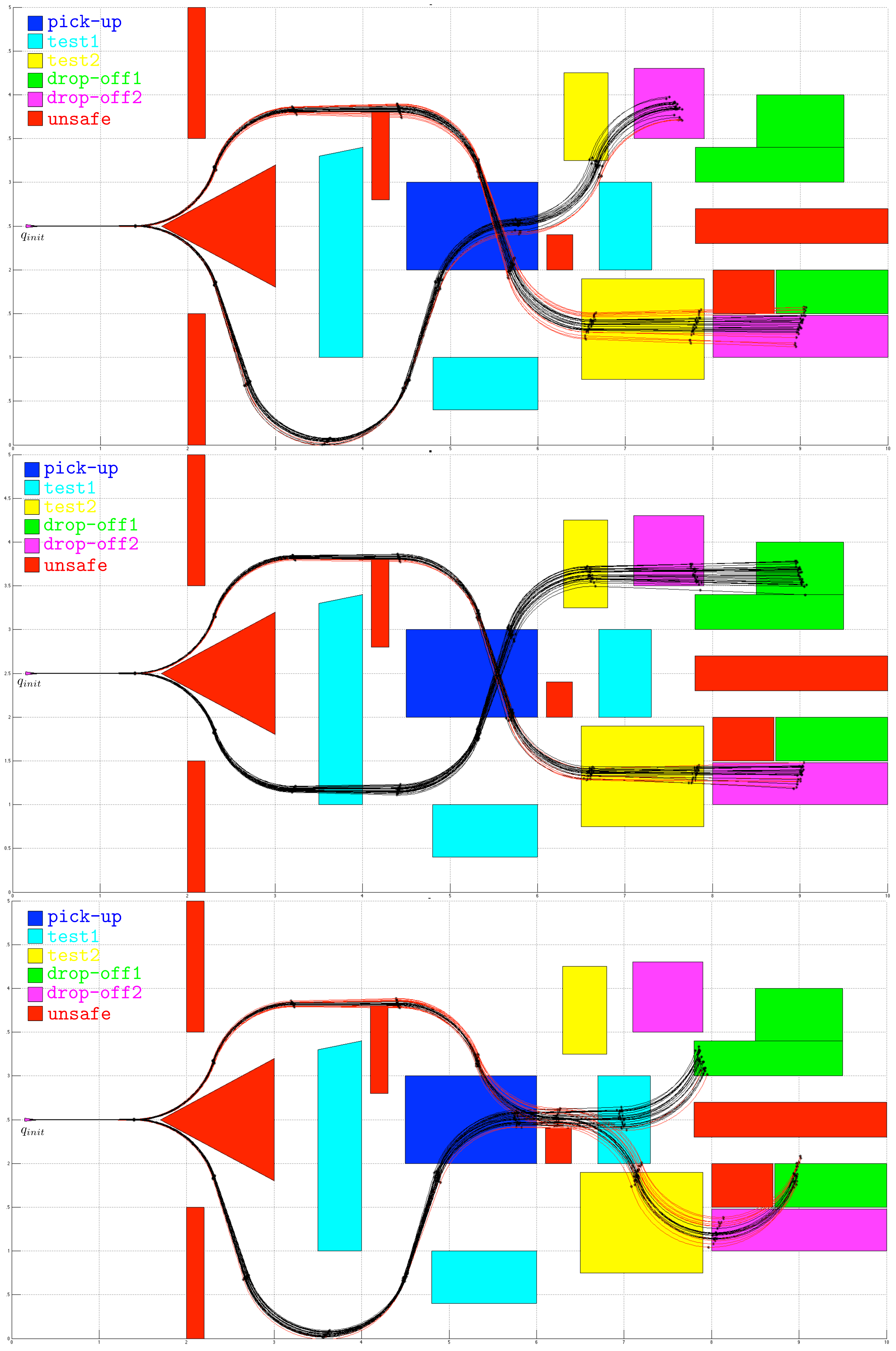}
\end{center}
\caption{50 sample state (position) trajectories for cases $A$, $B$ and $C$ (to be read top to bottom) obtained by simulating the original system under the corresponding vehicle control strategies. Satisfying and violating trajectories are shown in black and red, respectively.  
}
\label{simulation results}
\end{figure}

\section{Conclusion and Future Work} \label{conclusion and future work}
We developed a human-supervised control synthesis method for a stochastic Dubins vehicle such that the probability of satisfying a specification given as formula in a fragment of PCTL over a set of environmental properties is maximized.  We modeled the uncertain motion of the vehicle in the environment as an MDP. For the PCTL fragment we introduced the specification update rules that guarantee the increase (or decrease) in the satisfaction probability. The specification can be updated, using the rules, until the supervisor is satisfied with both the updated specification and the corresponding satisfaction probability. We introduced two efficient algorithms for synthesizing MDP control policies, one from an initial PCTL formula and another from an updated PCTL formula. Both algorithms exploit the special structure of the MDP, as well as the structure of the PCTL formula. The second algorithm produces an updated solution by reusing the initial solution. We proposed an offline and an online application of this method. 

Future work includes extensions of this work to controlling different types of vehicle models, allowing for richer temporal logic specifications, and experimental validations.

\bibliographystyle{alpha} 
\bibliography{references}   

\end{document}